\documentclass[twoside]{article}
\pdfoutput=1
\usepackage[preprint]{aistats2022}
\usepackage{amsmath,amsthm,amssymb}
\usepackage{authblk}
\usepackage{mathtools}
\usepackage{algorithm,algorithmic}
\usepackage{booktabs} 
\usepackage{hyperref}
\usepackage{url}
\usepackage{footnote}
\usepackage{graphicx}
\usepackage{natbib}
\usepackage{subfig}
\usepackage{bm}
\usepackage{paralist}
\usepackage{calrsfs}
\usepackage{subfiles}
\usepackage{breqn}

\theoremstyle{plain}
\newtheorem{theorem}{Theorem}
\newtheorem{corollary}{Corollary}[theorem]
\newtheorem{lemma}[theorem]{Lemma}

\newtheorem{claim}{Claim}
\newtheorem{definition}{Definition}

\newtheorem*{assumption}{Assumption}
\newtheorem*{lemma*}{Lemma}

\newcommand{\bb}[1]{\mathbb{#1}}

\newcommand{\wh}[1]{\widehat{#1}}
\newcommand{\bfa}[1]{\boldsymbol{#1}}
\newcommand{\fk}[1]{\mathfrak{#1}}
\DeclareMathAlphabet{\pazocal}{OMS}{zplm}{m}{n}
\newcommand{\ca}[1]{\pazocal{#1}}

\newcommand{\pa}[1]{\mathcal{#1}}

\newcommand{\pij}{\wh\pi(i),\wh\pi(j)}

\newcommand{\bbfa}[1]{\bar{\bfa #1}}

\newcommand{\whp}[1]{\wh\pi(#1)}

\newcommand{\oml}{M_{{i_1},l}}
\newcommand{\tml}{M_{{i_2},l}}

\newcommand{\delt}{\Delta_{i_1,i_2,l}}

\DeclareMathOperator*{\argmax}{arg\,max}

\begin{document}

%

%

\twocolumn[

\aistatstitle{Achievability and Impossibility of Exact Pairwise Ranking}

\aistatsauthor{ Yihan He  }

\aistatsaddress{ Courant Institute, NYU \\ \href{mailto:yihan.he@nyu.edu}{yihan.he@nyu.edu}} ]

\begin{abstract}
We consider the problem of recovering the rank of a set of $n$ items based on noisy pairwise comparisons. We assume the SST class as the family of generative models.  Our analysis gave sharp information theoretic upper and lower bound for the exact requirement, which matches exactly in the parametric limit. Our tight analysis on the algorithm induced by the moment method gave better constant in Minimax optimal rate than ~\citet{shah2017simple} and contribute to their open problem. The strategy we used in this work to obtain information theoretic bounds is based on combinatorial arguments and is of independent interest.
\end{abstract}
\section{Introduction}
Pairwise ranking focuses on the task of deciding the rank within an unordered set with $n$ items. The decision process is built upon the observation over the outcomes of comparison between every two parties in multi-round fashion. This problem stems from the identification of the best player in the tournament and the estimation of relative rank among teams. In the more recent literature, a variety of novel applications, including recommender systems and peer grading, applies the algorithms  for this problem. Motivated by these active examples, we study the information theoretic and computational limits of a general class of parametric models (e.g., the SST class ) in this problem.

In the problem of interest, we consider a collection of $n$ parties and assume that the observation consists of outcomes on the multiple rounds of pairwise comparisons. We assume the outcome is Bernoulli, where party $i$ wins over $j$ with probability $M_{i,j}\in(0,1)$ that relies on the relative rank between them and the outcome across different pairs to be independent of each other. Alike ~\citet{shah2017simple}, we also considered the random design observation model, where the observation of any pairs in each round is observed with probability $p$. In particular, we focus on the problem of exact recovery, where the estimator converges almost surely to the ground truth as the scale of the problem goes to infinity. Under this requirement, we formally discuss the fundamental limits in this problem. A typical question in this problem is \textit{ When do there exists estimators that guarantee weak consistency ?} and \textit{ When do there exists no estimator that guarantees strong consistency ?}

We note that the impossibility and achievability for exact recovery are closely related to the \textit{All or Nothing} phenomenon. This phenomenon is fundamental and is a direct result of the Borel-Cantelli lemmas. It states that when the number of correlated or weakly correlated random variables goes to infinity, the probability of events concerning a countable union of them will be either $1$ (e.g., All) or $0$ (e.g., Nothing). However, to decide the probability being $0$ or $1$ is nontrivial, which involve the estimation of boundary conditions. The information theoretic impossibility and achievability establish the threshold of this phase transition. Generally, by impossibility, we mean that no estimator exists while achievability argues for its converse. Another notion termed computational achievability is also used where the estimator is assumed to be constructed out of polynomial complexity procedure.

The major contribution in this work can be concluded in two folds. First, we established the sharp information theoretic limits that converge exactly in parametric limits. This result comes from the idea of finding the combinatorial events that result in the failure and is necessary in the success of MAP estimator, which yield a sharp estimate on the limits. ~\citet{shah2017simple} has established the Minimax rate for this problem whereas our work is the Bayes rate under uniform prior. Our result is not implied by theirs and is definitely stronger, since Bayes rate is strictly upper bounded by Minimax rate. Secondly, we gave an improved efficient algorithm with polynomial complexity. This yields a better rate and, in the meanwhile, yields a better constant between computational achievability and its converse, which has also been posted as an open problem in ~\citet{shah2017simple}.

\textbf{Organization}: We organize this work through multiple sections. Section 2 introduces the general literature of this problem. Section 3 introduces the mathematical formulation of the problem, with extra background knowledge for information theoretic limits. Section 4 gives the formal result of our analysis together with the proofs. Section 5 gives sharp guarantees of moment method.

\section{Related Work}
Many historical and recent works have made remarkable progress in the many forms of this problem. The problem was first studied in the parametric models like BTL ~\citep{bradley1952rank} and Thursture~\citep{thurstone1927law} by ~\citet{shah2015estimation}. Then the SST class is proposed~\citep{shah2016stochastically} as the general class of models for the ranking problem. In particular, the problem of identification of top $k$ items are studied extensively in ~\citet{shah2016stochastically,shah2017simple}, where they gave Minimax optimal rate as well as an efficient algorithm for the ranking in SST class. ~\citet{chatterjee2015matrix} studied the problem of estimating the probability matrix itself.~\citet{rajkumar2014statistical,rajkumar2015ranking} gave several result on the pairwise ranking problem under different models. Their result guarantees the consistency of the counting algorithm, but the result is loose compared with ~\citet{shah2017simple}. Our work further strengthened the bound via a simple but effective analysis of the moment method.

Several works also considered the top-$k$ recovery problem, which is a weaker form than the full recovery problem considered in this work.
~\citet{chen2017competitive} give an efficient algorithm for the top $K$ ranking algorithm in the SST class.~\citet{chen2015spectral} gives a spectral MLE algorithm that can achieve exact ranking with high probability as regularity conditions. However, their parametric model is BTL, which is shown to be not as effective in characterizing the real tournament result. ~\citet{shah2017simple} considered both the full recovery and top $k$ recovery problem and proved that the counting algorithm is Minimax optimal. Instead of focusing on the Minimax rate, we considered the Bayes rate, which is stated as the limit over the prior. This is a harder problem as the standard theory suggests Bayes risk is upper bounded by Minimax risk and they almost never converge. ~\citet{heckel2019active} studies the active ranking, where their result suggests that a logarithmic factor can be dropped from the sample complexity by switching from passive learning regime to active learning regime. We stick to the multi-round complete observation model.

\section{Problem Formulations}
Here we give a formal statement of the problem. Given an integer $n\geq 2$, consider the collections of $n$ parties, indexed by $[n]:=\{1,...,n\}$ each with their own 'quality' value $w_1,...,w_n\in\bb R^+$. Since the ranking is agnostic of the order in the generative model, without loss of generality, we can assume that their quality is monotonically decreasing with indices. e.g. $w_1 >\ldots> w_n$.  Let $M_{i,j}\in(0,1)$ be the probability that the $i$-th strongest party wins over the $j$-th in a single round of competition.

A model in the SST class is defined by a strictly increasing function $F$ with $\sup(F)\in[0,1]$ such that
\begin{dmath*}
    M_{i,j}=F(w_i-w_j)
\end{dmath*}
In this work, we denote $\gamma_{max}=\sup_{i,j} M_{i,j}$ and $\gamma_{min}=\inf_{i,j} M_{i,j}$. We denote $\bfa M(\ca P)=\bfa P(\bfa A|\ca P)$ by the induced measure for random matrix $\bfa A\in\bb A=\{0,1\}^{n\times n}$. $\ca P$ is the state space on which prior is defined. We let $\bb P:[n]\rightarrow[n]$ be the symmetric group of permutations on $[n]$. Let $\Pi\in\bb P$ be the random variables defined by $\bfa P(\Pi=\pi)=\fk P(\pi)$ for all $\pi\in\bb P$, where $\fk P$ is the prior, which we assumed to be uniform. The ordered outcome graphs are written as $G(\ca V,\bfa A)$, where $\ca V$ is the finite set of vertices and $\bfa A$ is the adjacency matrix. Moreover, we say a graph $G$ is induced by some $\bfa A$ if its adjacency matrix is $\bfa A$. 


The problem is stated as follows: We assume that first $\Pi$ is drawn from $\bb P$ according to $\fk P$, without loss of generality, let $\Pi=\pi^*$. The generative model permutes the index of parties according to $\pi^*$ and generate a series of total $m$ observations, denoted by $\bfa A_1^m=\{\bfa A^1,...,\bfa A^m\}$ that is i.i.d. drawn from $\bb A$ according to $\bfa M(\Pi=\pi^*)$. We denote the normalized ensemble by $\ca A^m=\frac{1}{m}\sum_{i=1}^{m}\bfa A^i$. The objective is to estimate $\pi^*$ based on $\bfa A_1^m$. Moreover, we denote $\bfa M(\Pi=\pi^*)$ by $\bfa M_{\pi^*}$ for simplicity. Denote $\bfa M_{\pi^*,i,j}=Bernoulli(M_{(\pi^*)^{-1}(i),(\pi^*)^{-1}(j)})$ be the Bernoulli measure parameterized by the ground truth probability of ground truth $i$ wins over $j$. The notations used throughout this work is that we denote random variables by capital letters and their value by lower-case ones.

\subsection{Model Specification}\label{gener}
 We consider the random design of experiments. Without loss of generality, we use $\bfa A$ to denote $\bfa A^i$ that is assumed to be of identical probability measure. We studied the random design matrix, assuming that with probability $1-p$ two parties will cancel their comparison. This model is also studied in ~\citet{shah2017simple}, which can be stated as follows
\begin{assumption}
For any two parties $i,j$ such that $i< j$, the random design gives
\begin{align*}
    &\bfa A_{i,j}\sim\bfa M_{\pi^*,i,j},\bfa A_{i,j}=-\bfa A_{j,i} &&\text{with probability }p\\
   &\bfa A_{j,i}=\bfa A_{i,j}=0 &&\text{with probability }1-p
\end{align*} 
\end{assumption}

In the sequel, we omit $\Pi=\pi^*$ and use $\bfa P(\cdot)$ to denote $\bfa P(\cdot|\Pi=\pi^*)$ where no confusions are made.

\subsection{Prelinimaries}
We review some classical notations throughout this work, which follow the standard ones in the network inference problems. The following definition of \textit{disagreement} is analogous to the Hamming distance  in this problem, defined by 
\begin{definition}[Disagreement]
The disagreement between two permutations $\pi_1,\pi_2\in\bb P$ is defined by
\begin{equation*}
    R(\pi_1,\pi_2)=\frac{1}{n}\sum_{i=1}^n\bfa 1_{\pi_1(i)\neq\pi_2(i)}
\end{equation*}

\end{definition}

Based on the consistency of estimators, we give definitions on the exact ranking:
\begin{definition}
We say the estimator $\wh\pi$ satisfies exact ranking if the following requirement is achieved:
\begin{equation*}
    \bfa P_{\bfa M,\fk P}(R(\wh\pi,\pi^*)=0)=1-o(1)
\end{equation*}
where $\pi^*$ is the ground truth. This also reminds us of the strong consistency or convergence a.e.
\end{definition}
Then the standard information theoretical limits are listed for completeness
\begin{definition}[Impossibility]
Assuming the parametric space is $\Theta$, we say the exact pairwise ranking is impossible in $\Theta$ if no estimator achieves exact ranking for any $\theta\in\Theta$.
\end{definition}
Similarly, contrary to the \textit{impossibility}, we also have \textit{achievablility} in both the information theoretic and computational sense.
\begin{definition}[Achievability]
Assuming the parametric space is $\Theta$, we say exact pairwise ranking is (information theoretically) achievable if for any $\theta\in\Theta$, there is an estimator $\wh\theta$ that satisfies exact ranking. Moreover, we say that exact pairwise ranking is computationally achievable if the procedure to obtain $\wh\theta$ is with polynomial complexity.
\end{definition}
The gap between these two types of achievabilities results in an information-computational gap. Intuitively, the impossibility and achievability stay at the two ends of phase transition. The former indicates uniform failure in the probability family, while the latter indicates uniform success. In simpler problems like detection with binary stochastic block model (SBM), the number of parameters is small, and these two notations converge. However, in a general problem, the two thresholds are not the same. The gap between the two thresholds governs the width of phase transition and will be an essential characteristic in our problem.

Then the following claim establishes a sufficient condition of impossibility.
\begin{claim}[Sufficiency of Impossibility]
When the parameter space $\Theta$ is finite, and the Bayes estimator fails exact ranking, the exact ranking is impossible. In particular, when the loss of the Bayes risk is $0/1$ loss, the Bayes estimator can be achieved by maximizing a posteriori. 
\end{claim}
By its definition, the Bayes estimator is optimal w.r.t. the prior on the disagreement metric. Hence, any estimator will have a higher probability of nonzero disagreement than the Bayes estimator, which implies that the Bayes estimator is on the limit of impossibility.

Previous literature ~\citep{shah2017simple} considers the exact ranking with Minimax optimality. However, Bayes estimator is equivalent to Minimax estimator when only a single estimator exists. Their results can be seen as an attempt to close the threshold of achievability and its converse, which differs from the problem studied here. It is also important to note that in this problem, the impossibility is not a converse argument for achievability, we discussed this phenomena in the next section. Hence, the contribution is largely exclusive.


\section{Fundamental Limits of Ranking}
The threshold staying between sharp impossibility and achievability resulted in the \textit{phase transition} phenomena. In this section, we discuss the two ends of it. Our result shows that using the condition over $\Theta(N)$ parameters, two sides of the threshold are closed at the limit. 


\subsection{Information Theoretic Lower Bound}
In this subsection, we present the discussion over the impossibility requirement. The method used here originates in multiple hypothesis testing. Recall that, for each vertex $i\in[n]$, the state space contains all possible outcomes of comparisons with the rest $n-1$ parties. And we can see this problem as trying to distinguish the order between any two parties, whose scale goes in $O(n^2)$. Intuitively, if two elements are too close to each other, the estimation will fail due to the impossibility of doing so. Although a similar argument is formed in ~\citet{shah2017simple}, who discuss the testing over $n-1$ adjacent pairs of parties to obtain Minimax lower bound, our discussion is different as we are focusing on lower bounding the risk of the test problem itself. This estimate hence provides a lower bound for the original problem. ~\citet{shah2017simple} considers $\bar\Delta:=\inf_{i\in[n-1]}\frac{1}{n}\sum_{k=1}^n(M_{i,k}-M_{i+1,k})$ as an indicator of success for the top-k discovery. Our result suggests that this indicator will be loose for the information theoretic threshold. Instead, we resort to $\frac{1}{n}\inf_{i\in[n-1]}\sum_{l\in[n]}(M_{i,l}-M_{i+1,l})^2$ as the indicator

Here we state the following sufficient conditions on the impossibility of exact ranking, which is analogous to the idea of \textit{Signal to Noise Ratio} (SNR) in signal processing. When the SNR is too low, the noise will abuse the original signal, making it impossible to be recovered. The following two conditions are sufficient for the failure.
\begin{compactenum}
\item \textbf{Connectivity:} If the undirected random graph induced by $\ca A^m$ is unconnected with $\Theta(1)$ probability, the exact ranking fails
\item \textbf{Failure of the MAP Estimator:} If the MAP estimator failed with probability $\Theta(1)$, then the exact ranking fails
\end{compactenum}
\subsubsection{Connectivity}
In the seminal work by ~\citet{erdos1960evolution}. The connectivity of a single Erdős–Rényi graph $G(n,p)$ satisfies that when $p=\frac{\log n+c}{n}$
\begin{align*}
    \bfa P(G(n,p) \text{ is connected })\rightarrow e^{-e^{-c}}
\end{align*}
Then we analogously establish the result for the ensembled random graphs $G^m(n,p)$ induced by $\ca A^m$
\begin{theorem}
When $p=\frac{\log n+c}{mn}$$$\bfa P(G^m(n,p)\text{ is connected })\rightarrow e^{-e^{-c}}$$
\end{theorem}
\begin{proof}
The proof is straightforward once we observe that $G^m(n,p)$ is the same as $G(n,1-(1-p)^m)$
\end{proof}
And the above theorem directly implies the following:
\begin{corollary}
When $c>1$ and $p\geq \frac{c\log n}{mn}$, then the graph $G^m(n,p)$ is connected.
\end{corollary}
In the sequel, we only considered the situation where the graph is connected, where $p>\frac{\log n}{mn}$
\subsubsection{Failure of MAP Estimator}
Here we discuss an estimate on the upper bound of exact ranking. The result we established is stated as

\begin{theorem}[Impossibility]
Assume that $M_{i,j}=\Theta(1)$ as $n\rightarrow\infty$. Let $\Delta_{i,j,l}=M_{i,l}-M_{j,l}$. Let $
    K_0:=\frac{1}{\gamma_{\max}}+\frac{1}{\gamma_{\min}}
$. Assume that $p>\frac{\log n}{mn}$. Then the following implies impossibility of exact ranking:
\begin{align*}
    \inf_{i,j}\sum_{l\in[n]\setminus\{i,j\}}\Delta^2_{i,j,l}\leq\frac{4\log n}{K_0mp}
\end{align*}
or, under a stricter condition
\begin{align*}
  \bar\Delta\leq\frac{2}{n}\sqrt{\frac{\log n}{K_0mp}}
\end{align*}
\end{theorem}

First, we decouple the random design matrix into a pointwise product between a Bernoulli filter with the original random matrix. Without loss of generality, we can always denote $\pi^*$ by the identity mapping since the real order in the random matrix does not affect the procedure of MAP.

We define the random design matrix by $\bar{\bfa A}_1^m=\{\bbfa A^1,\ldots,\bbfa A^m\}$ and decouple the generative procedure. Assume that we have a Bernoulli random matrix $\bfa Q\in[0,1]^{n^2}$ defined by
\begin{align*}
    \bfa Q_{i,j}=\left\{\begin{matrix}
1 & \text{with probability } p \\
0 &  \text{with probability } 1-p\\
\end{matrix}\right.
\end{align*}
subjecting to $\bfa Q_{i,j} =\bfa Q_{j,i}$. Then we can decompose
$    \bbfa A=\bfa A\otimes \bfa Q$
where $\otimes$ is the element-wise product. Moreover, we have that for any $a>0$ :
\begin{align*}
\bfa P(\bbfa A_{i,j}=a)&=\bfa P(\bfa A_{i,j}=a,\bfa Q_{i,j}>0)\\
&=(1-p)\bfa P(\bfa A_{i,j}=a)
\end{align*}
and $\bfa P(\bbfa A_{i,j}=0)=1-p$. We decouple the entry sum of $\bfa A^m_1$ into positive and negative part:
\begin{align*}
    \ca A_{i,j}^{m,+}&=\sum_{k=1}^m\bfa 1_{\bfa Q_{i,j}=1}\bfa 1_{\bfa A_{i,j}^k=1}\\
    \ca A_{i,j}^{m,-}&=\sum_{k=1}^m\bfa 1_{\bfa Q_{i,j}=1}\bfa 1_{\bfa A_{i,j}^k=-1}
\end{align*}

And we rewrite the maximum a posteriori criteria as
\begin{align*}
    \wh\pi_{\text{MAP}}&=\argmax_{\wh\pi\in\bb P}\bfa P(\Pi=\wh \pi|M,\bbfa A_1^m)\\
                     &=\argmax_{\wh\pi\in\bb P}\prod_{k=1}^m\bfa P(\bbfa A^k|M,p,\Pi=\wh\pi)\bfa P(\Pi=\wh\pi)\\
                     &=\argmax_{\wh\pi\in\bb P}\sum_{k=1}^m\log(\bfa P(\bbfa A^k|M, p,\Pi=\wh \pi))\\
                     &=\argmax_{\wh\pi\in\bb P}\sum_{k=1}^m\sum_{i< j}^n\log(\bfa P(\bbfa A_{\wh \pi(i),\wh \pi(j)}|M_{ i, j},p,\Pi=\wh\pi))\\
                     &=\argmax_{\wh\pi\in\bb P}\sum_{k=1}^m\sum_{\substack{i<j:\\\bfa Q_{\pij}>0}}\\
                     &\qquad\qquad\log(\bfa P(\bfa A_{\wh \pi(i),\wh \pi(j)}|M_{ i, j},p,\Pi=\wh\pi))
\end{align*}
According to the likelihood function of Binomial random variables, we have
\begin{align*}
    \log(\bfa P(\bfa A_{i,j}|M_{i,j}))&=\bfa A_{i,j}\cdot\log(M_{i,j})\\
    &+(1-\bfa A_{i,j})\cdot\log(1-M_{i,j})
\end{align*}

Then we look into the objective that we maximize, recall that we denote the ground truth $\pi^*$ to be the identity bijection. We define $F:\bb N\times\bb N\rightarrow\bb R$ as the score function
\begin{align*}
   & F(i,j)=\sum_{k=1}^m\sum_{l\in [n]}\bfa 1_{\bfa Q_{\wh\pi(i),l}=1}\log(\bfa P(\bfa A_{\wh\pi(i),l}|M_{j,l},\Pi=\wh\pi))\\
    &=\sum_{l\in[n]\setminus\{{i_1},{i_2}\}}\ca A_{\whp i,l}^{m,+}\log(M_{ j,l})+\ca A_{\whp i,l}^{m,-}\log(1-M_{ j,l})
\end{align*}Hence, terms in the log likelihood function concerning $\wh\pi(i_1)$ and $\wh\pi(i_2)$ are termed by $F(i_1,i_1) $ and $F(i_2,i_2)$.

We can quickly identify an event that will cause the MAP procedure fails to return correct estimator: $$F(i_1,i_2)+F(i_2,i_1)\leq F(i_1,i_1)+F(i_2,i_2)$$, given by
\begin{align*}
&\sum_{l\in[n]\setminus\{{i_1},{i_2}\}}\bigg(\ca A_{\whp {i_1},l}^{m,+}\log(M_{ i_2,l})+\ca A_{\whp {i_2},l}^{m,+}\log(M_{ i_1,l})\\&+\ca A_{\whp {i_1},l}^{m,-}\log(1-M_{ {i_2},l}) +\ca A_{\whp {i_2},l}^{m,-}\log(1-M_{ {i_1},l}) \bigg)\\
&\geq\sum_{l\in[n]\setminus\{{i_1},{i_2}\}}\bigg(\ca A_{\whp {i_1},l}^{m,+}\log(M_{ i_1,l})+\ca A_{\whp {i_2},l}^{m,-}\log(M_{ i_2,l})\\&+\ca A_{\whp {i_1},l}^{m,-}\log(1-M_{ {i_1},l}) +\ca A_{\whp {i_2},l}^{m,-}\log(1-M_{ {i_2},l})\bigg)
\end{align*}
which implies that
\begin{align*}
    E_{i_1,i_2}&:=\sum_{l\in[n]\setminus\{{i_1},{i_2}\}}(\ca A_{\whp {i_1},l}^{m,+}-\ca A_{\whp {i_2},l}^{m,+})\log\bigg(\frac{M_{{i_2},l}}{M_{{i_1},l}}\bigg)\\
    &+(\ca A_{\whp {i_1},l}^{m,-}-\ca A_{\whp {i_2},l}^{m,-})\log\bigg(\frac{1-M_{{i_2},l}}{1-M_{{i_1},l}}\bigg)\geq 0
\end{align*}
The idea to approach the upper bound for MAP estimator is to explore over the event such that MAP fails, namely $\bfa P(E_{i_1,i_2}\geq 0)$.

To approach this problem, we construct a set of random variables $\bfa C^{(i_1,i_2)}\in\bb R^{2\times n}$ such that when $l\notin\{i_1,i_2\}$,
\begin{align*}
\bfa C^{(i_1,i_2)}_{i_1,l}=\left\{\begin{matrix}
\log\big(\frac{M_{{i_2},l}}{M_{{i_1},l}}\big)& \text{with probability } M_{{i_1},l}p \\
\log\big(\frac{1-M_{{i_2},l}}{1-M_{{i_1},l}}\big) &  \text{with probability } (1-M_{{i_1},l})p\\
0& \text{ with probability } 1-p
\end{matrix}\right.
\end{align*}
and 
\begin{align*}
  \bfa C^{(i_1,i_2)}_{i_2,l}=\left\{\begin{matrix}
\log\big(\frac{M_{{i_1},l}}{M_{{i_2},l}}\big)& \text{with probability } M_{{i_2},l}p \\
\log\big(\frac{1-M_{{i_1},l}}{1-M_{{i_2},l}}\big) &  \text{with probability } (1-M_{{i_2},l})p\\
0& \text{ with probability } 1-p
\end{matrix}\right.
\end{align*}
Hence, the error probability can be rewritten through sampling $m$ independent copies of $\bfa C^{(i_1.i_2)}$, denote by $\{\bfa C^{(1)},\ldots,\bfa C^{(m)}\}$, and we have
\begin{align*}
    \bfa P(E_{i_1,i_2}\geq 0)=\bfa P\big (\sum_{s\in[m]}\sum_{i\in\{i_1,i_2\}}\sum_{l\in[n]\setminus\{{i_1},{i_2}\}}\bfa C^{(s)}_{i,l}\geq 0\big )
\end{align*}
To approach this transformed version of problem, we start by reviewing a few technical lemmas.

\begin{lemma}[Paley-Zygmund]
For random variable $Z>0$, for $\theta\in(0,1)$ we have
\begin{align*}
    \bfa P(Z\geq\theta\bb E[Z])\geq(1-\theta)^2\frac{\bb E[Z]^2}{\bb E[Z^2]}
\end{align*}
\end{lemma}

which directly implies
\begin{corollary}[Second Order Method]
$
\bfa P(Z\geq 0)\geq\frac{\bb E[Z]^2}{\bb E[Z^2]}
$
\end{corollary}
The following lemma gives a sufficient condition of impossibility. The idea in proving this lemma comes from Borel-Cantelli, which can be proved by the above corollary.
\begin{lemma}
Assume that $M_{i,j}=\Theta(1)$ as $n\rightarrow\infty$ (e.g. non-vanishing) and $|\sup_i (M_{i,i+1}-M_{i-1,i})|=\Delta \rightarrow 0$ as $n\rightarrow\infty$. If $\bfa P(E_{i,i+1}>0)=\Omega(\frac{1}{n})$ for all $i\in[n-1]$, then the exact ranking will be impossible.
\end{lemma}
The above lemma argues for an sufficient condition of impossibility. Hence if we can argue for the condition to be asymptotically almost surely (a.a.s.) , then we can obtain a bound of impossibility. This leads to the formal proof of impossibility
\begin{proof}
The proof follows from Paley-Zygmund inequality.
Let $\theta=\frac{1}{\bb E[\exp(tZ)]}$, we have:
\begin{align*}
    \bfa P(Z\geq 0)&=\bfa P(\exp(Z)\geq 1)\\
    &\geq \bigg (1-\frac{1}{\bb E[\exp(tZ)]}\bigg)^2\frac{\bb E[\exp(tZ)]^2}{\bb E[\exp(2tZ)]}
\end{align*}
For the simplicity of notation, we denote
\begin{align*}
    \pa A_l=\frac{\tml}{\oml},\;\;\;\;\;\pa B_l=\frac{1-\tml}{1-\oml}
\end{align*}
And we note that $\log \pa A_l\cdot\log \pa B_l<0$. Without loss of generality, we assume $\log \pa A_l>0$, $\log\pa B_l<0$ and let $\bfa X_l\sim Bernoulli(\oml)$, $\bfa Y_l\sim Bernoulli(\tml)$ in what follows.

The moment generating function for random variable $\bfa C_{i_1,l}^{(i_1,i_2)}$ by
\begin{align*}
    \psi_{\bfa C^{(i_1,i_2)}_{i_1,l}}(t)&=\bb E[\exp(t\bfa C^{(i_1,i_2)}_{i_1,l})]
    \\&=\bigg(\frac{\tml}{\oml}\bigg)^tp\oml+1-p\\
    &+\bigg(\frac{1-\tml}{1-\oml}\bigg)^tp(1-\oml)\\
    &=\pa A_l^tp \oml+1-p+\pa B_l^tp(1-\oml)\\
    &=1-p+pD_{f_t}(\bfa X_l\Vert\bfa Y_l)
\end{align*}
where for $\bfa X_l\sim P$, $\bfa Y_l\sim Q$, we have that $D_{f_t}(\bfa X_l\Vert\bfa Y_l)=D_f(P\Vert Q)=\int (\frac{dP}{dQ})^tQ(dx)$ is the $f$-divergence with $f_t(x)=x^t$.

Similarly, for the random variable $\bfa C_{i_2,l}$, we have
\begin{align*}
    \psi_{\bfa C^{(i_1,i_2)}_{i_2,l}}(t)&=\bigg(\frac{\oml}{\tml}\bigg)^tp\tml+1-p\\
    &+\bigg(\frac{1-\oml}{1-\tml}\bigg)^tp(1-\tml)\\
    &=\pa A_l^tp \tml+1-p+\pa B_l^tp(1-\tml)\\
    &=1-p+pD_{f_t}(\bfa Y_l\Vert\bfa X_l)
\end{align*}

Assembling pieces, we have
\begin{align*}
    \bb E [\exp( tE_{i_1,i_2} )]&=\bb E[\exp (\sum_{i\in\{1,2\}}\sum_{l\in[n]\setminus\{{i_1},{i_2}\}}t\bfa C_{i,l})]\\
    &=\prod_{i\in\{1,2\}}\prod_{l\in[n]\setminus\{{i_1},{i_2}\}} \bb E[\exp(t\bfa C_{i,l})]
\end{align*}
which implies
\begin{align*}
    &\bfa P(E_{i_1,i_2}> 0)>\frac{(\bb E[\exp(tE_{i_1,i_2})]-1)^2}{\bb E[\exp(2tE_{i_1,i_2})]}\\
    &>\frac{\big (\prod_{s\in[m]}\prod_{i\in\{1,2\}}\prod_{l\in[n]\setminus\{{i_1},{i_2}\}}\bb E[\exp(t\bfa C^{(s)}_{i,l})]-1\big )^2}{\prod_{s\in[m]}\prod_{i\in\{1,2\}}\prod_{l\in[n]\setminus\{{i_1},{i_2}\}}\bb E[\exp(2t\bfa C^{(s)}_{i,l})]}
\end{align*}
From Lemma 4, we note that impossibility is implied by
\begin{align*}
    \bfa P(E_{i_1,i_2}>0)=\Omega \bigg(\frac{1}{n}\bigg)
\end{align*}
for all $i_2-i_1=1$. This is implied by
\begin{align*}
    &\big(\prod_{s\in[m]}\prod_{i\in\{1,2\}}\prod_{l\in[n]\setminus\{{i_1},{i_2}\}}\bb E[\exp(t\bfa C_{i,l})]-1\big)^2\\
    &=\Omega\bigg(\frac{1}{n}\prod_{s\in[m]}\prod_{i\in\{1,2\}}\prod_{l\in[n]\setminus\{{i_1},{i_2}\}}\bb E[\exp(2t\bfa C_{i,l})]\bigg)
\end{align*}
taking the logarithm over both side, we have
\begin{align*}
    &\sum_{s\in[m]}\sum_{i\in\{1,2\}}\sum_{l\in[n]\setminus\{{i_1},{i_2}\}}(2\log(\bb E[\exp(t\bfa C_{i,l}^{(s)})]-1)\\
    &\qquad\qquad\qquad\qquad-\log\bb E[\exp(2t\bfa C^{(s)}_{i,l})])\geq -\log n
\end{align*}
Note that $\bb E[\exp(t\bfa C_{i_1,i_2})]\rightarrow 1$ as $n\rightarrow\infty$, the above can be implied by
\begin{align*}
    \frac{1}{n}\sum_{l\in[n]\setminus\{{i_1},{i_2}\}}2\log(\bb E[\exp(t\bfa C_{i_1,l}+t\bfa C_{i_2,l})-1)&\\
    -\log\bb E[\exp(2t\bfa C_{i_1,l}+2t\bfa C_{i_2,l})]\leq\frac{\log n}{nm}&
\end{align*}
By the definition of $f_t$-divergence, let $\Delta_{i_1,i_2,l}=\oml-\tml$, we have
\begin{align*}
    &D_{f_t}(\bfa X_l\Vert\bfa Y_l)= \bigg(\frac{\tml}{\oml}\bigg)^t\oml+\bigg(\frac{1-\tml}{1-\oml}\bigg)^t(1-\oml)\\
    &=(\oml)^{-(t-1)}(\oml+\Delta_{i_1,i_2,l})^t\\
    &+(1-\oml)^{-(t-1)}(1-\oml-\Delta_{i_1,i_2,l})^t\\
    &= (\oml)^{-(t-1)}\\
    &\cdot(\oml^t+t\Delta_{i_1,i_2,l}\oml^{t-1}+\frac{1}{2}t(t-1)\Delta_{i_1,i_2,l}^2\oml^{t-2})\\
    &+(1-\oml)^{-(t-1)}[(1-\oml)^t-t\Delta_{i_1,i_2,l}(1-\oml)^{t-1}\\
    &+\frac{1}{2}t(t-1)\Delta_{i_1,i_2,l}^2(1-\oml)^{t-2})+o(\Delta_{i_1,i_2,l}^2)\\
    &=1+\frac{1}{2}t(t-1)(\oml^{-1}+(1-\oml)^{-1})\Delta_{i_1,i_2,l}^2\\
    &+ o(\Delta_{i_1,i_2,l}^2)
\end{align*}
Collecting pieces, we concluded impossibility can be implied by
\begin{align*}
    \frac{1}{n}\sum_{l\in[n]\setminus\{{i_1},{i_2}\}}(-2t(2t-1))(\frac{1}{\oml}+\frac{1}{1-\oml})\Delta^2_{i_1,i_2,l}&\\
    \leq \frac{\log n}{nm p}&
\end{align*}
Let $t=\frac{1}{4}$ and note that
\begin{align*}
    K_0:=\frac{1}{\gamma_{\max}}+\frac{1}{\gamma_{\min}}
\end{align*}
we finally conclude by
\begin{align*}
    \frac{K_0}{4}\sum_{l\in[n]\setminus\{{i_1},{i_2}\}}\Delta_{i_1,i_2,l}^2\leq\frac{\log n}{mp}
\end{align*}
which completes the proof.
\end{proof}

\subsection{Information Theoretic Upper Bound}
Then we discuss an estimate on the lower bound of exact ranking. The result we established as the information theoretic achievability is stated as
\begin{theorem}
Assume that the probability is non-vaninshing, Let $\Delta_{i,j,l}=M_{i,l}-M_{j,l}$, then the following implies that there exists algorithm that rank exactly.
\begin{align*}
    \inf_{i,j}\sum_{l\in[n]}\Delta_{i,j,l}^2\geq\frac{K_0\log n}{4mp}
\end{align*}
Or under stricter condition:
\begin{align*}
    \bar\Delta\geq\sqrt{\frac{K_0\log n}{4nmp}}
\end{align*}
\end{theorem}
To obtain the information theoretic upper bound, we rely on the following lemma.
\begin{lemma}[Information Theoretic Achievability]
 If MAP estimator does not recover the ground truth, then with $\Theta (1)$ probability there exists $i_1,i_2\in[n]$ such that 
 \begin{align*}
     F(i_1,i_2)+F(i_2,i_1)\leq F(i_1,i_1)+F(i_2,i_2)
 \end{align*}
\end{lemma}

By the above lemma, we found out a necessary event of success. If this event is asymptotically impossible, then we can argue for the success of MAP estimator, which finally establish the achievability. We then give formal proof of achievability.
\begin{proof}
Let $\bfa X_{i,l}\sim Bernoulli(\bfa M_{i,l})$. We recall that the centered moment generating function of $\bfa C_{i_1,l}^{(i_1,i_2)}+\bfa C_{i_2,l}^{(i_1,i_2)}$ is
\begin{align*}
   & \psi_{\bfa C_{i_1,l}^{(i_1,i_2)}+\bfa C_{i_2,l}^{(i_1,i_2)}}(t)\\
   &=\bb E[\exp(\bfa C_{i_1,l}^{(i_1,i_2)}+\bfa C_{i_2,l}^{(i_1,i_2)}-\bb E[\bfa C_{i_1,l}^{(i_1,i_2)}+\bfa C_{i_2,l}^{(i_1,i_2)}])]\\
    &=\bigg(\frac{M_{i_2,l}(1-M_{i_1,l})}{M_{i_1,l}(1-M_{i_2,l})}\bigg)^{tp(M_{i_2,l}-M_{i_1,l})}\\
    &\cdot(1-p+pD_{f_t}(\bfa X_{i_1,l}\Vert\bfa X_{i_2,l})\cdot(1-p+pD_{f_t}(\bfa X_{i_2,l}\Vert\bfa X_{i_1,l}))
\end{align*}
taking the logarithm over both side, we have
\begin{align*}
    &\log\big(\psi_{\bfa C_{i_1,l}^{(i_1,i_2)}+\bfa C_{i_2,l}^{(i_1,i_2)}}(t)\big)\\
    &\leq\frac{t^2\delt^2p}{2}\bigg (\frac{1}{M_{i_1,l}}+\frac{1}{1-M_{i_1,l}}+\frac{1}{M_{i_2,l}}+\frac{1}{1-M_{i_2,l}}\bigg )\\
    &\qquad\qquad\qquad\qquad\qquad\qquad\qquad\qquad+o(\delt^2)\\
    &\leq p\delt^2t^2\bigg(\frac{1}{\gamma_{\max}}+\frac{1}{\gamma_{\min}}\bigg)+o(\delt^2)
\end{align*}
By Bennet's inequality, we have for $x\geq 0$, we have
\begin{align*}
    &\bfa P\bigg (\sum_{s\in[m]}\sum_{l\in[n]\setminus\{i_1,i_2\}}\bfa C_{i_1,l}^{(s)}+\bfa C_{i_2,l}^{(s)}-\bb E[\bfa C_{i_1,l}^{(s)}+\bfa C_{i_2,l}^{(s)}]\geq x\bigg )\\
    &\leq\frac{\prod_{l\in[n]\setminus\{i_1,i_2\}}\psi_{\bfa C_{i_1,l}^{(i_1,i_2)}+\bfa C_{i_2,l}^{(i_1,i_2)}}(t)}{\exp (tx)}
\end{align*}
Let $K_0=\frac{1}{\gamma_{\max}}+\frac{1}{\gamma_{\min}}$.
Minimizing the R.H.S. we have
\begin{align*}
   &\bfa P\bigg (\sum_{s\in[m]}\sum_{l\in[n]\setminus\{i_1,i_2\}}\bfa C_{i_1,l}^{(s)}+\bfa C_{i_2,l}^{(s)}-\bb E[\bfa C_{i_1,l}^{(s)}+\bfa C_{i_2,l}^{(s)}]\geq x\bigg )\\
   &\leq\exp\bigg(-\frac{x^2}{2pm\sum_{l\in[n]\setminus\{i_1,i_2\}} \delt^2K_0}\bigg)
\end{align*}
which implies that
\begin{align*}
    &\bfa P(E_{i_1,i_2}\geq 0)\\
    &=\bfa P\bigg (\sum_{s\in[m]}\sum_{l\in[n]\setminus\{i_1,i_2\}}\bfa C_{i_1,l}^{(s)}+\bfa C_{i_2,l}^{(s)}-\bb E[\bfa C_{i_1,l}^{(s)}+\bfa C_{i_2,l}^{(s)}]\\
    &\geq -m\sum_{l\in[n]\setminus\{i_1,i_2\}}\bb E[\bfa C_{i_1,l}^{(i_1,i_2)}+\bfa C_{i_2,l}^{(i_1,i_2)}]\bigg )\\
    &\leq\exp\bigg(-\frac{m^2(\sum_{l\in[n]\setminus\{i_1,i_2\}}\bb E[\bfa C_{i_1,l}^{(i_1,i_2)}+\bfa C_{i_2,l}^{(i_1,i_2)}])^2}{2pm\sum_{l\in[n]\setminus\{i_1,i_2\}} \delt^2K_0}\bigg)\\
    &\leq\exp\bigg(-\frac{8mp\sum_{l\in[n]\setminus\{i_1,i_2\}} \delt^2}{K_0}\bigg)
\end{align*}
where we applied the fact that when $\Delta_{i_1,i_2}\rightarrow 0$ and note that when $M_{i_1,l}$ and $M_{i_2,l}$ are on the same side of $\frac{1}{2}$
\begin{align*}
    \bb E[\bfa C_{i_1,l}^{(i_1,i_2)}+\bfa C_{i_2,l}^{(i_1,i_2)}]&=p(M_{i_2,l}-M_{i_1,l})\\
    &\qquad\log\bigg(\frac{M_{i_2,l}(1-M_{i_1,l})}{M_{i_1,l}(1-M_{i_2,l})}\bigg)\\
    &\geq 4\delt^2p
\end{align*}
    We upperbound the joint event by the union bound:
    \begin{align*}
        &\bfa P(\bigcup_{i_1,i_2\in[n]}E_{i_1,i_2})\leq\sum_{i_1,i_2\in[n]:i_1\neq i_2}\bfa P(E_{i_1,i_2})\\
        &\leq\frac{n(n-1)}{2}\exp\bigg(-\frac{8mp\inf_{i,j}\sum_{l\in[n]\setminus\{i,j\}}\Delta_{i,j,l}^2}{K_0}\bigg)
    \end{align*}
    To guarantee the consistency of MAP, we will need:
    \begin{align*}
        \inf_{i,j}\sum_{l\in[n]\setminus\{i,j\}}\Delta^2_{i,j,l}\geq\frac{K_0\log n}{4mp}
    \end{align*}
    \textbf{Discussion:} The result given in those two bounds exactly equal at the limit of $\gamma_{\min}=\gamma_{\max}$. Recall that in this case $K_0=4$. The reason that those two bounds are not converging in the general case (e.g. $\gamma_{max}>\frac{1}{2}$) is the concavity of information theoretic achievable region in the space spanned by $\Theta(n^2)$ random variable. When only $\Theta(n)$ random variables are considered, the gap in the bound is not sharp in general.
\end{proof}

\section{Efficient Algorithm}
In this section, we present the guarantee for moment method, which is efficient and has a finer guarantee than the analysis of counting algorithm by ~\citet{shah2017simple}. In particular, our algorithm recovers rank based on the estimated first moment of row sum in the ensembled random matrix. Our result is formally stated as:
\begin{theorem}
The proposed efficient algorithm achieves exact recovery when 
\begin{align*}
    \bar\Delta\geq\sqrt\frac{\log n}{mnp}
\end{align*}
\end{theorem}


To obtain this result, we first state the moment method. The first moment estimate of entries of probability matrix is given by
\begin{align*}
    \wh M_{i,j}=\frac{\ca A_{i,j}^m}{2p}+\frac{1}{2}=\frac{1}{2pm}\sum_{l\in[m]}\bfa A_{i,j}^l+\frac{1}{2}
\end{align*}
with $
    \wh M_{i,i}=\frac{1}{2}
$
And it is easy to see that the unbiasedness
\begin{align*}
    \bb E[\wh M_{i,j}]=M_{i,j}
\end{align*}
Although $p$ is explicitly included here, the estimation can be carried out when $p$ is unknown a prior, since the goal is ranking instead of estimation. In that case, we will be estimating $pM_{i,j}$ instead.

Then the algorithm ranked the party based on the entry sum of estimator :
\begin{align*}
    \ca M_{i}=\sum_{j\in[n]}\frac{\wh M_{i,j}}{n}
\end{align*}
Here we introduced the sufficient and necessary conditions on the success of pairwise ranking with this method. 
\begin{claim}
Let $\wh G_{i,j}=\ca M_{i}-\ca M_{j}$. Then the exact recovery is equivalent to:
\begin{equation*}
    \wh G_{i,i+1}>0\;\;\text{ for all }i\in[n-1] \text{ a.a.s. }
\end{equation*}
\end{claim}
In the sequel, we also denote 
\begin{equation*}
    G_{i,j}=\frac{1}{n}\sum_{k=1}^n(M_{i,k}-M_{j,k})
\end{equation*}
as the popularity gap.

Based on the above claim, we can establish the guarantees for the moment methods. The technical details are delayed to the appendix.

Our method is within the tractable computational region because the sorting and the mean estimation can all be conducted in polynomial complexity.

\textbf{Discussion: }
Our upper bound gave a sharper rate than \citet{shah2017simple} and closed the constant from $8$ to $1$. 
\begin{compactenum}
\item The technique used in this analysis is Bernett's inequality and the first-order Taylor approximation to achieve sharpness. As Berstein's inequality is a direct result of Bernett's, our analysis yields a sharper bound than ~\citet{shah2017simple}.
\item The multiple hypothesis testing problems only engaged $\Theta(n)$ pairs while ~\citet{shah2017simple} engaged $\Theta(n^2)$ pairs. 
\end{compactenum}

On the other hand, this bound does not contradict what we have for the information theoretic upper bound, which follows that theorem 7 is a condition on the $1$-norm. If we state the sufficient condition in the form analogous to the information theoretic upper bound. It will become $\inf_{i,j}\sum_{l\in[n]}\Delta^2_{i,j,l}\geq\frac{n\log n}{mp}$, which is over pessimistic.
\section{Future Work}
The open problem that has not been addressed here is whether the constant gap can be closed at the minimax optimal rate. ~\citet{shah2017simple} gave a $\frac{1}{70}$ constant for this problem via Fano's method on multiple hypothesis testing constructing on a $\Theta(n)$ pairs. Hence, our central question will be whether a sharper rate can be established for the Minimax lower bound. 

The second problem is if we can relax the requirement to almost exact recovery. In this case, MAP is not admissible. ~\citet{shah2017simple} studied a notion called approximated recovery and gave guarantees for them under constant error gap in Hamming distance. However, their argument does not argue for vanishing error, and the estimator that works for the non-vanishing threshold does not imply almost exact recovery. Hence, it will be interesting to see if we can establish the impossibility condition on the almost exact recovery.
\section{Conclusion}
This work studies the problem of pairwise ranking under the requirement of exact ranking. We established the information theoretic upper and lower bound that is exact in the limit. We performed a tight analysis on the moment method for this problem and gave a sharper guarantee for the Minimax optimal rate. Open question on whether the lower bound can also be improved is raised.

\bibliographystyle{plainnat}
\bibliography{bib.bib}
\end{document}


%

%

\onecolumn
\aistatstitle{Instructions for Paper Submissions to AISTATS 2022: \\
Supplementary Materials}

\section{Proof of Lemma 4}
\begin{proof}
Denote event $D_{i_1,i_2}: E_{i_1,i_2}\geq 0$ and $\bb E[D_{i_1,i_2}]\bb E[D_{i_3,i_4}]\neq\bb E[D_{i_1,i_2}\cap D_{i_3,i_4}]$ by $(i_1,i_2)\overset{n}{\sim} (i_3,i_4)$ and the following quantities:
\begin{align*}
    \mu_n&:=\sum_{i_1\neq i_2}^{n}\bfa P( D_{i_1,i_2})\\
    \gamma_n&:=\sum_{(i_1,i_2)\overset{n}{\sim}(i_3,i_4)}\bfa P(D_{i_1,i_2}\cap D_{i_3,i_4})
\end{align*}
We denote $X_n:=\sum_{i_1\neq i_2}^n\bfa 1_{D_{i_1,i_2}}$. Note that
\begin{align*}
    Var[X_n]=\sum_{i_1\neq i_2}^n Var[\bfa 1_{D_{i_1,i_2}}]+\sum_{} Cov[\bfa 1_{D_{i_1,i_2}},\bfa 1_{D_{i_3,i_4}}]
\end{align*}
we have
\begin{align*}
    Var[\bfa 1_{D_{i_1.i_2}}]=\bb E[(\bfa 1_{D_{i_1,i_2}})^2]-(\bb E[\bfa 1_{D_{i_1,i_2}}])^2\leq\bfa P(D_{i_1,i_2})
\end{align*}
In particular, when $D_{i_1,i_2}$ and $D_{i_3,i_4}$ are independent
\begin{align*}
    Cov[\bfa 1_{D_{i_1,i_2}},\bfa 1_{D_{i_3,i_4}}]=0
\end{align*}
otherwise, if $(i_1,i_2)\overset{n}{\sim}(i_3,i_4)$,
\begin{align*}
    Cov[\bfa 1_{D_{i_1,i_2}}&,\bfa 1_{D_{i_3,i_4}}]\\
    &=\bb E[\bfa 1_{D_{i_1,i_2}}\bfa 1_{D_{i_3,i_4}}]
    -\bb E[\bfa 1_{D_{i_1,i_2}}]\bb E[\bfa 1_{D_{i_3,i_4}}]\\
    &=\bfa P(D_{i_1,i_2}\cap D_{i_3,i_4})-\bfa P(D_{i_1,i_2})\bfa P(D_{i_3,i_4})
\end{align*}
Now we can separate the event into two different cases:
\begin{compactenum}
\item $i_1,i_2$ and $i_3,i_4$ are not overlapping
\item $i_1,i_2$ and $i_3,i_4$ overlap by 1 index.
\end{compactenum}
\subsection{First Case}

Let $\bfa\Gamma=\sum_{s\in[m]}\bfa C^{(s)}_{i_1,i_3}+\bfa C^{(s)}_{i_1,i_4}+\bfa C^{(s)}_{i_2,i_3}+\bfa C^{(s)}_{i_2,i_4}$ be the sum of cross indices entries and note that $D_{i_1,i_2}\cap D_{i_3,i_4}$ can be rewritten according to the multiplicity of sub-indices by:
\begin{align*}
    &\bfa P(D_{i_1,i_2}\cap D_{i_3,i_4})-\bfa P(D_{i_1,i_2})\bfa P(D_{i_3,i_4})\\
    &=\int \bfa P\big(D_{i_1,i_2}\cap D_{i_3,i_4}|\bfa\Gamma=c\big)\cdot\bfa P\big(\bfa\Gamma=c\big)dc
    -\bfa P(D_{i_1,i_2})\bfa P(D_{i_3,i_4})\\
    &=\int_{E^n_\epsilon}\bfa P(E_{i_1,i_2}\geq-c|\bfa\Gamma=c)(\bfa P(E_{i_3,i_4}\geq -c|\bfa\Gamma=c)
    -\bfa P(E_{i_3,i_4}>0))\bfa P(\bfa\Gamma=c)dc
    \\&+\int_{\bb R\setminus E^n_\epsilon}\bfa P(E_{i_1,i_2}\geq-c|\bfa\Gamma=c)(\bfa P(E_{i_3,i_4}\geq -c|\bfa\Gamma=c)
    -\bfa P(E_{i_3,i_4}>0))\bfa P(\bfa\Gamma=c)dc
\end{align*}
Then we studied the middle term via the characteristic function. Note that as $n\to\infty$
\begin{align*}
    \bfa\Gamma\overset{p}{\to} 0
\end{align*}
For $\epsilon>0$, we construct $E_\epsilon^n=\{\omega:\bfa P(|\bfa\Gamma|\geq\omega)\geq\epsilon\}$ and have
\begin{align*}
    \limsup_n E_\epsilon^n=\{0\}
\end{align*}
Hence, by dominated convergence theorem, the second term goes to $0$. 
And, for the first term, consider the characteristic function
\begin{align*}
    \bb E[\exp(it\bfa\Gamma)]\to 0
\end{align*}
we have
\begin{align*}
    \bb E[\exp(it(E_{i_3,i_4}))|\bfa\Gamma]\rightarrow\bb E[\exp(itE_{i_3,i_4})]
\end{align*}
By Levy's theorem, we conclude that
\begin{align*}
    \lim_n\sup_{c\in E^n_\epsilon}|\bfa P(E_{i_3,i_4}\geq -c|\bfa\Gamma=c)-\bfa P(E_{i_3,i_4}\geq 0)|\to 0
\end{align*}
Using dominated convergence theorem again and we conclude that
\begin{align*}
    \bfa P(D_{i_1,i_2}\cap D_{i_3,i_4})\to\bfa P(D_{i_1,i_2})\bfa P(D_{i_3,i_4})
\end{align*}
\subsection{Second Case}

For the second case, we write $(i_3,i_4)$ as $(i_2,i_3)$ without loss of generality. Then, for any $i_1,i_2,i_3\in[n]$, we have
\begin{align*}
    \bfa P(D_{i_1,i_2}\cap D_{i_2,i_3})&\leq \min\{\bfa P(D_{i_1,i_2}), \bfa P(D_{i_2,i_3}\})\\
    &=O(n)\bfa P(D_{i_1,i_2})\cdot\bfa P(D_{i_2,i_3})
\end{align*}
and we conclude that
\begin{align*}
    \gamma_n=O(\mu_n^2)
\end{align*}
which implies that
\begin{align*}
    \frac{Var[X_n]}{(\bb E[X_n])^2}\leq\frac{\mu_n+\gamma_n}{\mu_n^2}=\frac{1}{\mu_n}+\frac{\gamma_n}{\mu_n^2}= O(1)
\end{align*}
Using the second moment method,
\begin{align*}
    \bfa P(X_n>0)&>\frac{(\bb EX_n)^2}{\bb E[X_n^2]}\\
    &=\frac{(\bb EX_n)^2}{(\bb EX_n)^2+Var[X_n]}\\
    &=\frac{1}{1+Var[X_n]/(\bb EX_n)^2}=\Omega(1)
\end{align*}
which implies impossibility.
\end{proof}

\section{Proof of Theorem 7}
\begin{proof}
The proof of this theorem relies on the tail analysis of $\wh G_{i,i+1}$ and the application of union bound.
Note that the following holds for all $l\in[m]$
\begin{align*}
    \psi_{M_{i,j}-\wh M_{i,j}}(\lambda)&=\bb E[\exp(\lambda M_{i,j}-\lambda \wh M_{i,j})]\\
    &= \exp(\lambda M_{i,j})\psi_{\sum_{l\in[m]}\frac{\bfa A_{i,j}^l}{2pm}+\frac{1}{2}}(\lambda)\\
    &=\exp(\lambda M_{i,j}-\frac{\lambda}{2})\bb E[\exp(-\sum_{l\in[m]}\frac{\lambda\bfa A_{i,j}^l}{2pm})]\\
    &=\exp(\lambda M_{i,j}-\frac{\lambda}{2})\prod_{l\in[m]}\bb E[\exp(-\frac{\lambda \bfa A_{i,j}^l}{2pm})]\\
    &=\exp (\lambda M_{i,j}-\frac{\lambda}{2} )\bigg[p M_{i,j}\exp\bigg(\frac{\lambda}{2mp}\bigg )+p(1-M_{i,j})\exp\bigg (\frac{-\lambda}{2mp}\bigg )+(1-p)\bigg]^m
\end{align*}
Henceforth, the centered m.g.f. for $\wh G$ is
\begin{align*}
    \psi_{ G_{i,j}-\wh G_{i,j}}(\lambda)&=\psi_{\frac{1}{n}\sum_{k=1}^n(M_{i,k}-\wh M_{i,k}-M_{j,k}+\wh M_{j,k})}(\lambda)\\
    &=\prod_{k=1}^n\psi_{M_{i,k}-\wh M_{i,k}}\bigg (\frac{1}{n}\lambda\bigg )\psi_{M_{j,k}-\wh M_{j,k}}\bigg (-\frac{1}{n}\lambda\bigg )\\
    &=\prod_{k\in[n]}\bb E[\exp(\lambda (M_{i,k}-\wh M_{i,k}))]\cdot\bb E[\exp(-\lambda ( M_{j,k}-\wh M_{j,k}))]\\
    &=\exp\bigg (\frac{\lambda}{n} \sum_{k\in[n]}(M_{i,k}-M_{j,k})\bigg )\\
    &\cdot\prod_{k=1}^n\bigg[pM_{i,k}\exp\bigg(\frac{\lambda}{2mnp}\bigg)+p(1-M_{i,k})\exp\bigg(\frac{-\lambda}{2mnp}\bigg)+1-p\bigg]^m\\
    &\cdot\prod_{k=1}^n\bigg[pM_{j,k}\exp\bigg(\frac{-\lambda}{2mnp}\bigg)+p(1-M_{i,k})\exp \bigg(\frac{\lambda}{2mnp}\bigg)+1-p\bigg]^m
\end{align*}
written in the logarithmic form, we have:
\begin{align*}
    \log\psi_{ G_{i,j}-\wh G_{i,j}}(\lambda)&=\frac{\lambda}{n}\sum_{k\in[n]}(M_{i,k}-M_{j,k})\\
    &+m\bigg(\sum_{k=1}^n\log\bigg(pM_{i,k}\exp\bigg(\frac{-\lambda}{2mnp}\bigg)+p(1-M_{i,k})\exp\bigg(\frac{\lambda}{2mnp}\bigg)+1-p\bigg)\bigg)\\
    &+m\bigg(\sum_{k=1}^n\log\bigg(pM_{j,k}\exp\bigg(\frac{\lambda}{2mnp}\bigg)+p(1-M_{i,k})\exp \bigg(\frac{-\lambda}{2mnp}\bigg)+1-p\bigg)\bigg)\\
    &= \frac{\lambda}{n}\sum_{k\in[n]}(M_{i,k}-M_{j,k})+\sum_{k\in[n]}\frac{M_{i,k}-M_{j,k}}{n}(-\lambda)+\frac{\lambda^2}{4mnp}\\
    &+\sum_{k\in[n]}\frac{\lambda^2}{8mn^2}((2M_{i,k}-1)^2-(2M_{j,k}-1)^2)+o(\lambda^2)
\end{align*}
Let \begin{align*}
    \kappa_{i,j}=\frac{1}{8mn^2}\sum_{k\in[n]}((2M_{i,k}-1)^2-(2M_{j,k}-1)^2)
\end{align*}
Then we obtain the conjugate of $\log\psi_{ G_{i,j}-\wh G_{i,j}}(\lambda)$:
\begin{align*}
    u_{i,j}(t)&=\sup_{\lambda}(\lambda t-\log\psi_{ G_{i,j}-\wh G_{i,j}}(\lambda))\\
    &=\sup_{\lambda}(\lambda t-\frac{\lambda^2}{4mnp}-\kappa_{i,j}\lambda^2)\\
    &=\frac{t^2}{4\kappa_{i,j}+\frac{1}{mnp}}
\end{align*}
And we apply Bernett's inequality:
\begin{equation*}
    \bfa P(G_{i,j}- \wh G_{i,j}\geq t)\leq\exp(-u(t))=\exp\bigg(-\frac{t^2}{4\kappa_{i,j}+\frac{1}{mnp}}\bigg)
\end{equation*}
Let $t=G_{i,j}=\frac{1}{n}\sum_{k=1}^n(M_{i,k}-M_{j,k})$, we concluded that:
\begin{align*}
    \bfa P(\wh G_{i,j}\leq 0)&\leq\exp\bigg(-\frac{G_{i,j}^2}{4\kappa_{i,j}+\frac{1}{mnp}}\bigg)\\
    &\leq\exp\bigg(-\frac{2mnpG_{i,j}^2}{2+\frac{p}{n}\sum_{k\in[n]}((2M_{i,k}-1)^2-(2M_{j,k}-1)^2)}\bigg)
\end{align*}

Taking the union bound over the set $\{(i,i+1):i\in[n-1]\}$, we concluded that:
\begin{align*}
    \bfa P\bigg (\bigcup_{i\in[n-1]}\{\wh G_{i,i+1}<0\}\bigg )&\leq \sum_{i=1}^{n-1}\bfa P(\wh G_{i,i+1}<0)\\
    &=\sum_{i=1}^{n-1}\exp\bigg(-\frac{2mnpG_{i,i+1}^2}{2+p\sum_{k\in[n]}((2M_{i,k}-1)^2-(2M_{i+1,k}-1)^2)}\bigg)\\
    &\leq n\sup_{i\in[n-1]}\exp\bigg(-\frac{mnpG_{i,i+1}^2}{1+4pG_{i,i+1}}\bigg)
\end{align*}
where we have used the fact that
\begin{align*}
    (2M_{i,k}-1)^2-(2M_{i+1,k}-1)^2=4(M_{i,k}-M_{i+1,k})(M_{i,k}+M_{i+1,k}-1)\leq 4(M_{i,k}-M_{i+1,k})
\end{align*}
Hence we will only need
\begin{align*}
    \inf_{i} G_{i,i+1}\geq\sqrt{\frac{\log n}{mnp}}
\end{align*}
\end{proof}